%% file: camera_ready_xaia_post_edits.tex
\documentclass{article}

\PassOptionsToPackage{numbers}{natbib}
\input{math_commands.tex}

\usepackage[final]{neurips_2023_final}

\usepackage[utf8]{inputenc} 
\usepackage[T1]{fontenc}    
\usepackage{hyperref}       
\usepackage{url}            
\usepackage{booktabs}       
\usepackage{amsfonts}       
\usepackage{nicefrac}       
\usepackage{microtype}      
\usepackage{xcolor}         
\usepackage{lmodern}

\usepackage{multirow}
\usepackage{multicol}   
\usepackage{color, colortbl, xcolor}
\usepackage{caption}

\def\arrvline{\hfil\kern\arraycolsep\vline\kern-\arraycolsep\hfilneg}

\usepackage{graphicx}
\usepackage{amsmath}
\usepackage{amsfonts}
\usepackage{amsthm}
\usepackage{wrapfig}
\usepackage{authblk}
\usepackage[symbol]{footmisc}

\theoremstyle{definition}
\newtheorem{theorem}{Definition}

\theoremstyle{definition}

\theoremstyle{definition}

\title{Sanity Checks Revisited: An Exploration to Repair the Model Parameter Randomisation Test} 

\author[ ]{
    Anna Hedström$^{1,3*}$
}
\author[ ]{
    Leander Weber$^{2,*}$
}
\author[ ]{
   Sebastian Lapuschkin$^{2,\dagger}$
}
\author[ ]{
    Marina Höhne$^{3,4,5\dagger}$
}

\affil[1]{Department of Electrical Engineering and Computer Science, TU Berlin}
\affil[2]{Department of Artificial Intelligence, Fraunhofer Heinrich Hertz Institute, Berlin, Germany}
\affil[3]{UMI Lab, Leibniz Institute of Agricultural Engineering and Bioeconomy e.V. (ATB)
}
\affil[4]{BIFOLD – Berlin Institute for the Foundations of Learning and Data}
\affil[5]{Department of Computer Science,
University of Potsdam}

\begin{document}

\maketitle
\def\thefootnote{*}\footnotetext{Equal Contribution}\def\thefootnote{\arabic{footnote}}

\def\thefootnote{\fnsymbol{footnote}}%
\setcounter{footnote}{2}
\footnotetext{Corresponding: \texttt{sebastian.lapuschkin@hhi.fraunhofer.de,MHoehne@atb-potsdam.de}}
\def\thefootnote{\arabic{footnote}}
\setcounter{footnote}{0}

\begin{abstract}
  The Model Parameter Randomisation Test (MPRT) is widely acknowledged in the eXplainable Artificial Intelligence (XAI) community
for its well-motivated evaluative principle: that the explanation function should be sensitive to changes in the parameters of the model function. However, recent works have identified several methodological caveats for the empirical interpretation of MPRT. 
To address these caveats, we 
introduce two adaptations to the original MPRT---\textit{Smooth MPRT} and \textit{Efficient MPRT}, where the former minimises the impact that noise has on the evaluation results through sampling and the latter circumvents the need for biased similarity measurements by re-interpreting the test through the explanation's rise in complexity, after full parameter randomisation. Our experimental results demonstrate that these proposed variants lead to improved metric reliability, thus enabling a more trustworthy application of XAI methods.

\end{abstract}

\section{Introduction}

The problem of evaluating the quality of an explanation method in eXplainable Artificial Intelligence (XAI) remains unsolved due to the frequent absence of ground truth explanation labels \cite{ Bellido1993,dasgupta2022, hedstrom2022quantus}. To address this issue, numerous quantitative evaluation methods have been proposed \cite{montavon2018, alvarezmelis2018robust, yeh2019, dasgupta2022, bhatt2020, alvarezmelis2018robust, agarwalstability, nguyen2020, bach2015pixel, samek2015, ancona2019, irof2020, yeh2019, rong2022, dasgupta2022, adebayo2018, sixt2019}.
Among these, the \textit{Model Parameter Randomisation Test} (MPRT) \cite{adebayo2018} measures the degree to which an explanation deteriorates as the model parameters are progressively randomised, i.e., layer-by-layer, starting from the output. This test posits that a greater difference in the explanation function's output, in response to parameter randomisation, signifies a higher quality of the explanation method.

In recent years, however, multiple independent research groups have provided constructive observations on the methodological choices in the original MPRT \cite{sundararajan2018, kokhlikyan2021investigating, yona2021, bindershort}. 
These encompass the selection of pairwise similarity for measuring explanation difference \cite{bindershort}, the order of layer randomisation \cite{bindershort}, the choice of explanation preprocessing \cite{sundararajan2018,bindershort} and the dependency on the model task \cite{yona2021, kokhlikyan2021investigating}. Given the considerable adoption of MPRT in the XAI community and its potential to influence the acceptance or rejection of different explanation methods, these raised concerns deserve a closer examination. 
Do evaluations based on the original MPRT optimally inform us about the quality of an explanation method, or can we improve 
its reliability?

\begin{figure}[!t]
    \centering
    \includegraphics[width=0.9\linewidth]{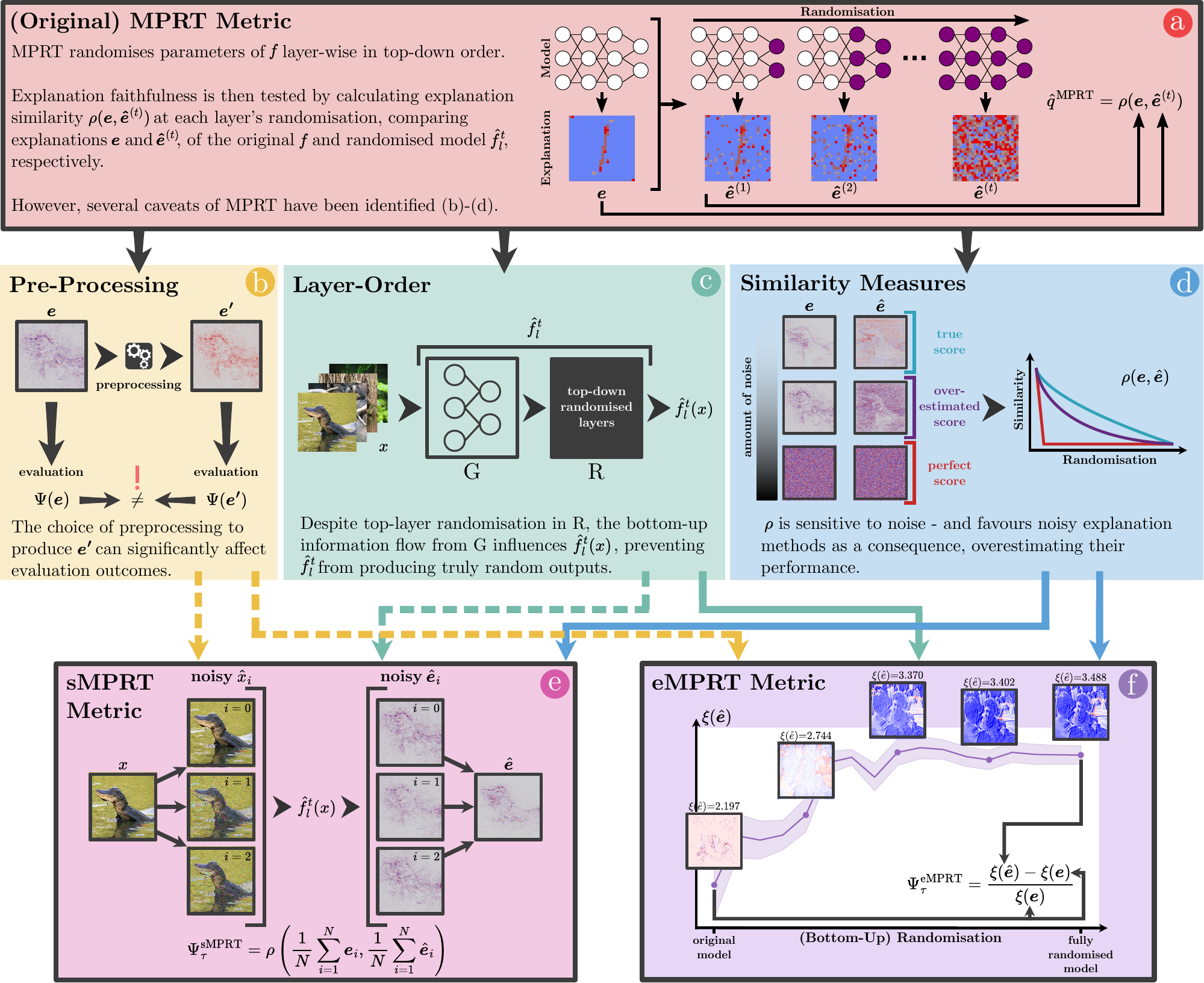}
    \captionsetup{font=footnotesize}
    \caption{
    Schematic visualisation of the original MPRT \cite{adebayo2018} (\emph{top}), identified shortcomings (\emph{middle}) and proposed solutions (\emph{bottom}). Solid arrows signify shortcomings directly addressed by our proposed metrics, while dashed arrows denote those addressed through ideas from previous work \cite{sundararajan2018,bindershort}. (a) The MPRT evaluates an explanation method by randomising $f$'s parameters in a top-down, layer-wise fashion and thereafter calculating explanation similarity $\rho(\ve, \hat{\ve})$ at each layer through comparing explanations $\ve$ of the original model $f$ and $\hat{\ve}$ of the randomised model $\hat{f}$. (b) \textit{Pre-processing}: normalisation and taking absolute attribution values significantly impact MPRT results, potentially deleting pertinent information about feature importance carried in the sign.
    (c) \textit{Layer-order}: top-down randomisation of layers in MPRT does not yield a fully random output, preserving properties of the unrandomised lower layers and thus affecting the evaluation of faithful explanations. (d) \textit{Similarity measures}: the pairwise similarity measures used in the original MPRT \cite{adebayo2018} are noise-sensitive, e.g., from gradient shattering and thus likely to impact evaluation rankings of XAI methods. (e) sMPRT extends MPRT by incorporating a ``denoising'' preprocessing step that averages explanations over $N$ perturbed inputs, aiming at reducing noise in local explanation methods.
    (f) eMPRT reinterprets MPRT by evaluating the faithfulness of the explanation method by comparing its rise in complexity, before and after full model randomisation. }
    \label{fig:overview}
    \vspace{-0.75em}
\end{figure}

The foundational idea behind MPRT---that the explanation function should be sensitive to the model's parameters---is both insightful and important. Our work seeks to build upon this initial work, addressing its shortcomings, namely, the order of layer randomisation and the selection of pairwise similarity measures, discussed in detail in Section \ref{issues} and illustrated in Figure \ref{fig:overview}.
We introduce two MPRT variants, \textit{Smooth MPRT} (sMPRT) and \textit{Efficient MPRT} (eMPRT), where the sMPRT mitigates sensitivity to shattering noise (cf. \cite{bindershort}) by denoising, i.e., averaging attributions over $N$ perturbed inputs and the eMPRT evaluates explanation faithfulness by quantifying the rise in complexity after full parameter randomisation, thus obviating a biased similarity measurement. We release these methods under the existing Quantus \cite{hedstrom2022quantus} evaluation framework\footnote{Code at: \url{https://github.com/understandable-machine-intelligence-lab/Quantus}.}.

Our methods, despite being an iteration, carry substantial importance. To deploy XAI in real-world applications, such as safety-critical domains, we need to focus on developing reliable quantitative evaluation metrics for existing explanation methods. 
This work contributes to ensuring that the continuous development of XAI methods remains both theoretically sound and empirically valid.

\section{Preliminaries}

For clarity in the following discussion, we first present the core notation.

\textbf{Local Explanations.} Let $f : \sX \mapsto \sY$, $f \in \sF$ be a differentiable black-box model that maps inputs $\vx \in \mathbb{R}^D$ to predictions $\vy \in \mathbb{R}^C$ with $C$ classes, where $\sF$ denotes the function space. We assume that $f$ consists of $L$ layers $f^1, \cdots, f^L$, such that $f(\vx) = f^{L} \circ \dots \circ f^{1}(\vx)$.  To interpret the behavior of $f$, \emph{local} explanation methods \cite{smilkov2017smoothgrad, ZeilerOcclusion, sundararajan2017axiomatic,selvaraju2019, bykov2021noisegrad} attribute importance scores to features of input $\vx$. Given a model's prediction $y$, we define an explanation ${\ve} \in \mathbb{R}^D$ via the explanation function $\Phi_{\lambda}: \mathbb{R}^D \times  \sF \times \sY \mapsto \mathbb{R}^D$, as $\ve = \Phi(\vx, f, y; \lambda)$. Then $\mathbb{E}$ denotes the space of possible explanations w.r.t. $f$ with $\Phi_{\lambda} \in \mathbb{E}$.

\textbf{Evaluating Explanations.}  Due to the lack of ground truth explanations for black-box models, evaluating the quality of $\Phi$ is generally non-trivial \cite{Bellido1993, bbox1997, dasgupta2022, hedstrom2023metaquantus}. This issue can be circumvented by instead measuring specific properties of explanations, such as stability (e.g. \cite{montavon2018, alvarezmelis2018robust, yeh2019}), complexity (e.g., \cite{nguyen2020, bhatt2020, chalasani2020}) or faithfulness (e.g., \cite{adebayo2018, bhatt2020, samek2015}). For a comprehensive overview of existing evaluation properties, we refer to \cite{hedstrom2022quantus} and \cite{hedstrom2023metaquantus}.
We define a quality estimate ${q} \in \mathbb{R}^M$ as the output of an evaluation function $\Psi_{\tau}: \mathbb{E} \times \mathbb{R}^{D} \times \sF \times \sY \mapsto \mathbb{R}$, parameterised by $\tau$ such that $q = \Psi_{\tau}(\Phi,\vx, f, y)$. 
Then $\sO$ denotes the space of possible evaluations w.r.t. $\ve$ and $\Psi \in \sO$.

\subsection{Model Parameter Randomisation Test (MPRT)}
\label{sec:mprt}

We define the original MPRT \cite{adebayo2018} in the following.
\begin{theorem}[MPRT]
\label{def-mprt}
\textit{Let $\Psi_{\tau}^{\text{MPRT}}: \mathbb{E} \times \mathbb{R}^{D} \times \mathbb{F} \times \mathbb{Y} \mapsto \mathbb{R}$ be an evaluation function that computes a quality estimate $\hat{q} \in \mathbb{R}$ that measures the similarity between the original explanation $\ve$ and the explanation $\hat{\ve} := \Phi(\vx, \hat{f}^{t}_l, y)$ corresponding to the perturbed model $\hat{f}^{t}_l$ randomised in a top-down fashion up to layer $l \in [L, L-1, \ldots, 1]$:} 
\begin{equation}\label{eq:q_mprt}
\hat{q}^{\text{MPRT}} = \rho(\ve, \hat{\ve}),
\end{equation}
\textit{with $\rho: \mathbb{R}^{D} \times \mathbb{R}^{D} \mapsto \mathbb{R}$ as a similarity function.}
\end{theorem}
For $\Phi$ to be sensitive to the parameter randomisation, Equation \ref{eq:q_mprt} should yield values such that $\hat{q}^{\text{MPRT}} \ll 1$. This means low similarity between $\ve$ and $\hat{\ve}$ for a given class $y$. 
As denoted in the superscript of $\hat{f}_l$, the direction of layer randomisation is completed top-down, i.e., $\hat{f}_l^{t}$ with $l \in [L, L-1, \ldots, 1]$. For bottom-up randomisation, we denote $\hat{f}_l^{b}$ with $l \in [1, 2, \ldots, L]$. 

\subsubsection{Methodological Caveats} \label{issues} 

In the following, we summarise the concerns that have been raised about various methodological aspects of 
MPRT, specifically its choice of (a) pre-processing, (b) layer-order and (c) similarity measure. For each issue, we put forward a 
solution.

\textbf{(a) Pre-processing.} In the original formulation of MPRT \cite{adebayo2018}, the explanations $\ve$ and $\hat{\ve}$ are normalised using their minimum and maximum values. This is problematic as these statistics are highly variable and almost arbitrary across attributions, complicating meaningful comparison across randomisations \cite{bindershort}. Taking absolute attribution values may erase meaningful information \cite{sundararajan2018}---a caveat already addressed by the seminal work \cite{adebayo2018}. \textit{(Proposed Solution) To maintain the original scale and distribution of the explanation, we recommend normalising $\ve$ by the  \textit{square root of the average second-moment estimate} \cite{bindershort} (see Equation \ref{eq:normalise-eq} in Supplement) which does not constrain attributions to a fixed range and enhances comparability across methods and randomisations \cite{bindershort}.}

\textbf{(b) Layer-order.} While one might intuitively expect a significant difference in the explanation output upon top-layer randomisation, i.e., $\rho(\ve, \hat{\ve}) \ll 1$ with $\hat{f}^{t}_l$ up to layer $l \in [L, L-1, \ldots, 1]$; \cite{bindershort} provides evidence to the contrary. As discussed in-depth in the original publication \cite{bindershort}, top-down randomisation induces only modest alterations in the forward pass where (i) irrelevant features from lower, non-randomised layers persist in higher, randomised layers, (ii) high activations in lower layers are relatively likely to continue to dominate the network's response and (iii) architectures with skip connections, e.g., ResNets  \cite{he2015deep} maintain a baseline explanation that stays constant even after randomisation, preserving certain features.
As such, only a limited degree of change between $\ve$ and $\hat{\ve}$ can be expected from faithful explanation methods post-model randomisation.
\textit{(Proposed Solution) To avoid preserving information in the forward pass, it is advisable to perform bottom-up randomisation (as further discussed in the Supplement \ref{sec:layerorder_supplement}) 
or perform the comparison of explanations only after full randomisation (see Equation \ref{eq:q_emprt}.}

\textbf{(c) Similarity Measures.} Another shortcoming identified by \cite{bindershort} is the property of \textit{SSIM} and other pairwise similarity measures, e.g., Spearman Rank Correlation, to be minimised by statistically uncorrelated random processes. This is problematic since it implies that certain explanation methods with  more intrinsic (shattering) noise, e.g., gradient-based methods, will be favoured in ranking comparisons, ultimately biasing evaluation outcomes. \textit{(Proposed Solution) While no solution has been proposed to date, we put forward the sMPRT and eMPRT modifications 
which separately addresses the issues by denoising attributions as a pre-processing step and replacing the pairwise similarity measure with a complexity measure, respectively.}

\section{Methods}

In the following, we provide mathematical details and motivations for sMPRT and eMPRT.

\subsection{Smooth Model Parameter Randomisation Test (sMPRT)}
\label{sec:smprt}

We first propose the sMPRT, which makes a small adaptation to the original MPRT, aimed at removing (shattering) noise observed in local explanation methods \cite{tessahan2022}. Inspired by \cite{smilkov2017smoothgrad,bykov2021noisegrad} we introduce a preprocessing step to the evaluation procedure---given an input $\vx$, generate $N$ perturbed instances $\hat{\vx}_i$, compute their attributions $\Phi(\hat{\vx}_i, f, y; \lambda)$ and then perform the MPRT evaluation on the averaged denoised attribution estimates.
We define the sMPRT in the following.
\begin{theorem}[sMPRT]
\label{def-smprt}
\textit{Let $\Psi_{\tau}^{\text{sMPRT}}: \mathbb{E} \times \mathbb{R}^{D} \times \mathbb{F} \times \mathbb{Y} \mapsto \mathbb{R}$ be an evaluation function that computes a quality estimate $\hat{q} \in \mathbb{R}$ that measures the similarity estimates between explanations $\ve_i := \Phi(\hat{\vx}_i, f, y; \lambda)$ and $\hat{\ve}_i := \Phi(\hat{\vx}_i, \hat{f}_l^b, y; \lambda)$ averaged over $i \in [1, N]$:}
\begin{equation}\label{eq:q_smprt}
\hat{q}^{\text{sMPRT}} = \rho\left(\frac{1}{N} \sum_{i=1}^N \ve_i, \frac{1}{N} \sum_{i=1}^N \hat{\ve}_i\right),
\end{equation}
\textit{where $\hat{\vx}_i = \vx + \eta_i$, $\eta_i \sim \mathcal{N}(0, \sigma)$ and $||\eta_i||_p \leq \epsilon$ holds with high probability, with $\sigma, \epsilon \in \mathbb{R}$.}
\end{theorem}
Similar to Equation \ref{eq:q_mprt}, to indicate low similarity between $\ve$ and $\hat{\ve}$, we expect values such that $\hat{q}^{\text{sMPRT}} \ll 1$ from Equation \ref{eq:q_smprt}. As discussed in the Supplement, to set $\sigma$, we follow the heuristic provided in the original publication \cite{smilkov2017smoothgrad}, i.e., $\sigma/(x_\text{max} - x_\text{min})$. For $N$, we find that $N=50$ provides consistent results (see Supplement \ref{sec:smprt_supplement}).

\textbf{Problems with sMPRT.} Despite sMPRT's capability to reduce sensitivity to shattering noise (see Figure \ref{fig:smprt}), sMPRT has notable drawbacks. First, it demands additional sampling ($N \geq 50$, see Figure \ref{fig:smprt_n}), significantly increasing computational cost compared to standard MPRT. 
Second, the practice of removing noise by adding noise is already a defining characteristic in some attribution methods, e.g., \textit{SmoothGrad} \cite{smilkov2017smoothgrad} and \textit{NoiseGrad} \cite{bykov2021noisegrad}, limiting sMPRT's efficacy with these methods and blurring the distinction with their baseline methods. Third, the degree of noisiness is an arguable property of attribution methods and removing it before evaluation may yield non-representative or biased results which may create performance discrepancies upon real-world application. Lastly, sMPRT introduces hyperparameters $\sigma$ and $N$, which may not be tunable on any given data domain, e.g., climate data \cite{bommer2023}. To this end, we next investigate alternative modifications to MPRT. 

\subsection{Efficient Model Parameter Randomisation Test (eMPRT)}
\label{sec:emprt}
To address caveats (b) and (c) and improve upon the efficiency of the original MPRT, we introduce eMPRT. This test effectively removes the layer-by-layer pairwise comparison between $\ve$ and $\hat{\ve}$ and instead compute the \textit{relative rise in explanation complexity} using only two model states, i.e., the original- and fully randomised model.
We define eMPRT in the following.
\begin{theorem}[eMPRT]
\label{def-emprt}
\textit{Let $\Psi_{\tau}^{\text{eMPRT}}: \mathbb{E} \times \mathbb{R}^{D} \times \mathbb{F} \times \mathbb{Y} \mapsto \mathbb{R}$ be an evaluation function that computes a quality estimate $\hat{q} \in \mathbb{R}$ that measures the relative rise in the complexity of the explanation from a fully randomised $\hat{f}$, i.e., $\hat{\ve} := \Phi(\vx, \hat{f}, y; \lambda)$, such that:}
\begin{equation}\label{eq:q_emprt}
\hat{q}^{\text{eMPRT}} = \frac{\xi(\hat{\ve}) - \xi(\ve)}{\xi(\ve)}
\end{equation}
\textit{with $\xi: \mathbb{R}^{D} \mapsto \mathbb{R}$ is a complexity function, e.g., defined through Equation \ref{eq:hist_entropy}.}
\end{theorem}
Since $\Phi$ ought to faithfully capture the behaviour of the model $f$---a fully randomised model (resulting in higher entropy of $f_L(\vx)$, as shown in Figure \ref{fig:emprt} (e) should therefore lead to reduced information content in its explanation $\hat{\ve}$. Equation \ref{eq:q_emprt} quantifies this heightened  explanation complexity as a rise in relative entropy, where a positive value, i.e., $\hat{q}^{\text{eMPRT}} > 0$ indicates a rise in complexity and a negative value, i.e., $\hat{q}^{\text{eMPRT}} < 0$ suggests a decrease in complexity. If $\hat{q}^{\text{eMPRT}} = 0$, no change in explanation complexity was observed.

To set $\xi$, we employ a histogram entropy measure, grounded in Shannon-Entropy \cite{Shannon48}. Specifically, attribution values $\ve$ are binned across $B$ distinct slots where for each $i^{\text{th}}$ bin we compute the frequency $c_i$ and derive the corresponding normalised probability, $p_i$ as follows:
\begin{equation} \label{eq:hist_entropy}
\xi(\ve) = -\sum_{i=1}^B p_i \log(p_i) \quad \text{where} \quad p_i = \frac{c_i}{\sum_{i=1}^B c_i}.
\end{equation}
Opting for this complexity measure 
offers three advantages: preservation of the inherent sign of attributions, implicit normalisation and adaptability to diverse dimensionalities and distributions, wherein $B$ can be contextually chosen.

\textbf{Advantages with eMPRT.} Generally, since explanations need only be computed twice, pre- and post-model randomisation, eMPRT is significantly more computationally efficient compared to MPRT. It further avoids issues related to layer-order as well as the choice of normalisation function (since the chosen entropy function already normalises implicitly, see Equation \ref{eq:hist_entropy}). 
Most importantly, by incorporating the initial unperturbed complexity estimate, i.e., $\xi(\ve)$ into Equation \ref{eq:q_emprt}, eMPRT anchors its quality estimate by the explanation method's inherent complexity, which makes the evaluation scores more comparable across explanation techniques and models. 

\section{Results}

The subsequent section provides a detailed analysis of our findings. 
Details on hyperparameters, as well as additional experiments, can be found in the Supplementary Material \ref{sec:experimental_setup_supplement}-\ref{sec:benchmarking-extra-results}. Code to replicate the experiments can be found in the GitHub repository\footnote{Code at: \url{https://github.com/annahedstroem/sanity-checks-revisited}}.

\begin{figure}[h]
\centering    \includegraphics[width=0.9\linewidth]{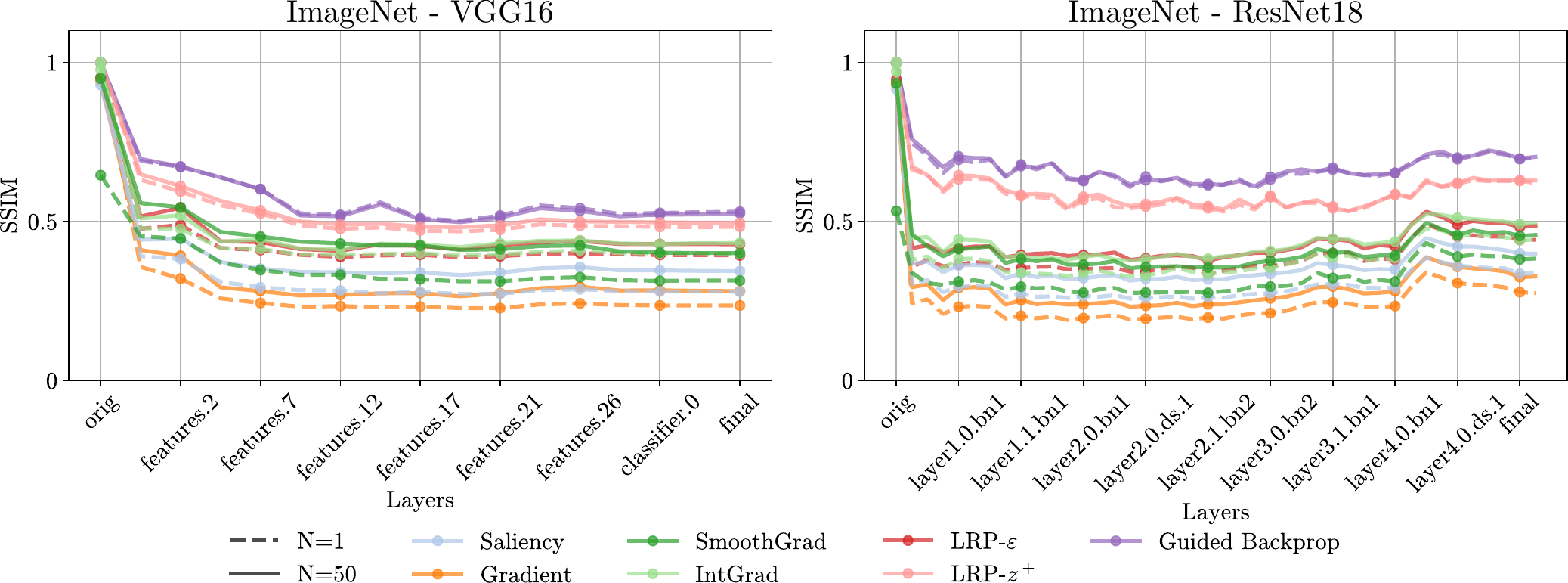}
    \captionsetup{font=footnotesize}
    \caption{sMPRT ($N=50$) versus MPRT ($N=1$) results, showing that denoising attributions with sMPRT can degrade \textit{SSIM} performance, which is most pronounced for gradient-based methods. \emph{orig} and \emph{final} denote the unrandomised and fully randomised model states, respectively.}
    \label{fig:smprt}
    \vspace{-0.5em}
\end{figure}

\textbf{Denoising Attributions with sMPRT.} Figure \ref{fig:smprt} visualises sMPRT scores with progressive randomisation of layers, in the bottom-up order. Results are shown for VGG-16 \cite{simonyan2014very} and ResNet-18  \cite{he2015deep} models for ImageNet \cite{ILSVRC15}, four different attribution methods and two choices of $N$ 
(where $N=1$ corresponds to the original MPRT). 
We observe that sMPRT---i.e., increasing the number of noise samples $N$ to 50---significantly affects the evaluation outcome, with the degree of change varying for different attribution methods. Interestingly, after removing noise by adding noise, the investigated attribution methods perform more similarly under randomisation. This observation complements the theoretical findings of \cite{bindershort}, which finds that the similarity scores computed by MPRT may be skewed in favour of (noisy) gradient-based explanation methods.

\begin{figure}[!t]
    \centering
    \includegraphics[width=0.9\linewidth]{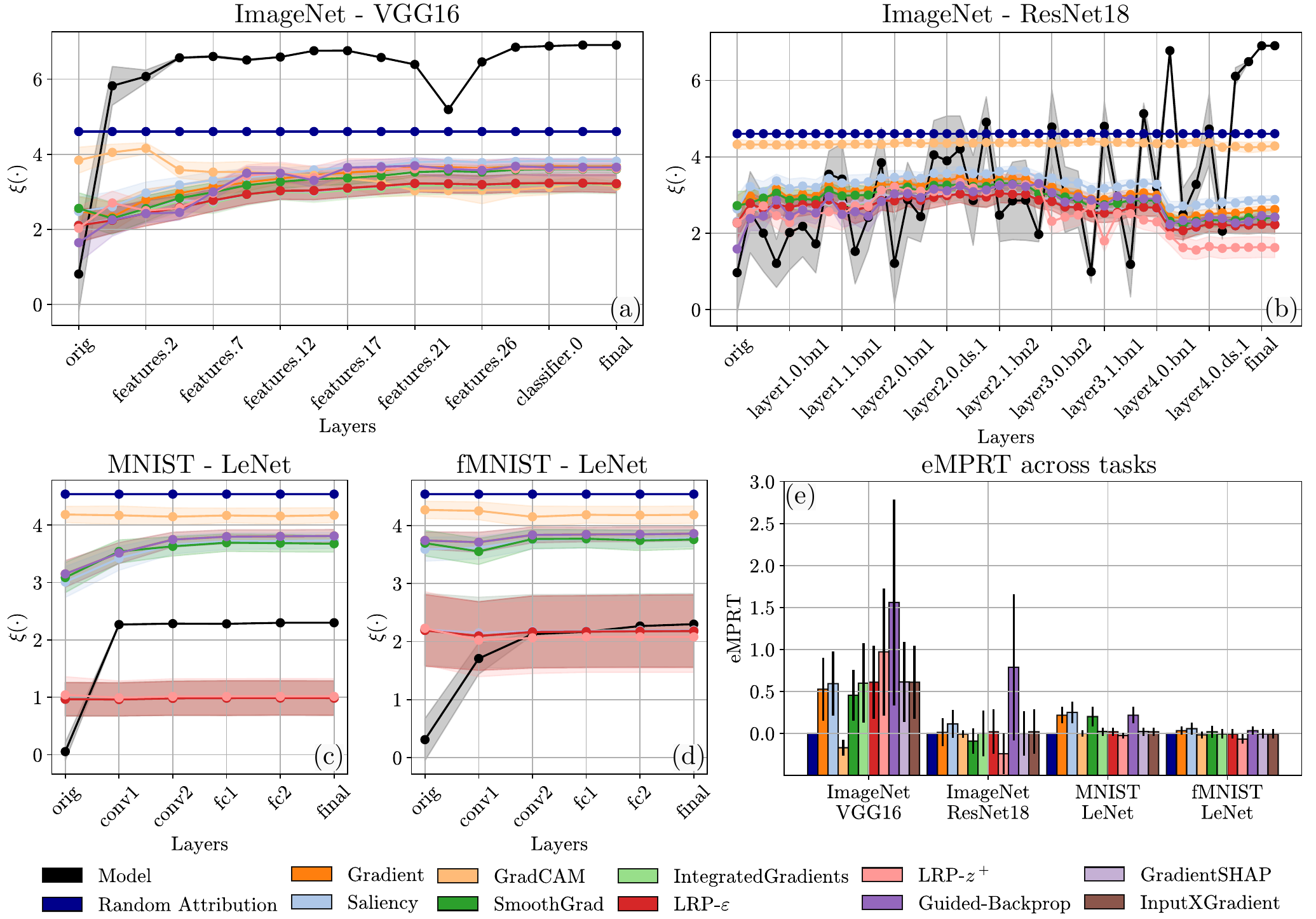}
    \captionsetup{font=footnotesize}
    \caption{eMPRT curves (a)-(d) depicting the rise in complexity $\xi(\hat{e})$ upon progressive layer randomisation with bottom-up randomisation $\hat{f}_l^b$ for different models and datasets and aggregated eMPRT scores (e) for comparative benchmarking of different explanation methods.}
    \label{fig:emprt}
    \vspace{-0.65em}
\end{figure}

\textbf{eMPRT in Action.} Figure \ref{fig:emprt} visualises eMPRT curves in (a)-(d) with progressive randomisation of layers 
and aggregated eMPRT scores in (e), across tested tasks. Results are shown for 
ten different attribution methods and a random attribution baseline.

First, the different explanation methods begin with varying complexity where no single explanation method consistently outperforms others across tested tasks. Even after full model randomisation, no method approaches the theoretical limit to randomness \emph{(dark blue)} one might expect from an explanation that acts maximally faithful to $\hat{f}$. Second,  
by comparing the model complexity (\emph{black}, computed as non-discretised entropy of the model's output layer post-softmax, see Equation \ref{shannon-entropy}) with the different explanation complexities, we may anchor our expectation of how the complexity of an ideal explanation, i.e., ``true to its model'' \cite{sturmfels2019}, ought to develop. Interestingly, none of the explanation functions mirror the model function's complexity. 
Third, the progression of eMPRT curves exhibits considerable variation across different tasks. While variation is expected (given architectural specifics such as ResNets' skip connections), it may be difficult for a practitioner to interpret the explanation method's faithfulness across the different layers.
As a more meaningful and robust measure of explanation faithfulness, we propose comparing explanations only before and after full model randomisation (see Equation \ref{eq:q_emprt}). In Supplement \ref{sec:emprt_supplement}, we further discuss the extent to which the evaluation rankings of explanation methods differ between eMPRT and MPRT. Notably, the random attribution method---serving as a theoretical lower bound for explanation faithfulness—--consistently receives lower evaluation scores under eMPRT compared to MPRT, signifying an advantage with the eMPRT metric.

\textbf{Meta-Evaluation.} Importantly, to understand how the variants MPRT, sMPRT and eMPRT differ in terms of metric performance characteristics, we set out a benchmarking experiment, following the meta-evaluation methodology outlined in \cite{hedstrom2022quantus}. 
Each metric is assigned a summarising meta-consistency score, i.e., $\overline{\text{MC}} \in [0, 1]$. A higher $\overline{\text{MC}}$ score, approaching 1, signifies greater reliability and superior performance based on the tested criteria.
Details on the meta-evaluation framework 
can be found in the original publication \cite{hedstrom2023metaquantus}. 
A more comprehensive breakdown of the benchmarking results is included in Supplement \ref{sec:benchmarking-extra-results}, with experimental details in \ref{sec:benchmarking_details}.

\begin{figure}[!t]
    \centering
    \includegraphics[width=0.95\linewidth]{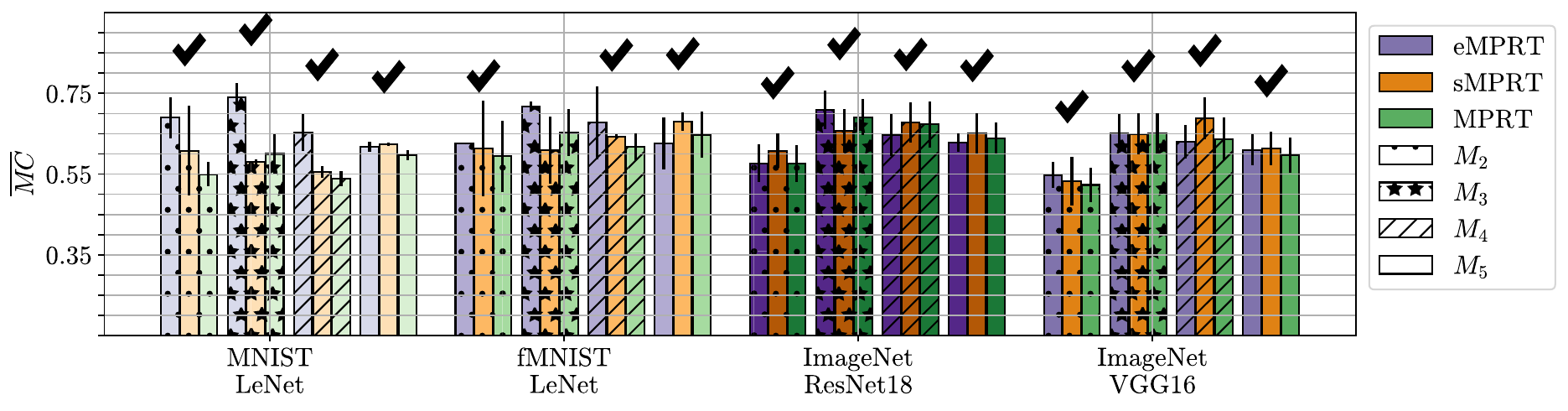}
    \captionsetup{font=footnotesize}
    \caption{Average $\overline{\text{MC}}$ scores per metric across tested tasks (see in Table \ref{table-benchmarking} in Supplement \ref{sec:benchmarking-extra-results}), aggregated over $3$ iterations with $K=5$ perturbations. Here, $M_2,$ $M_3,$ $M_4,$ and $M_5$ represent different combinations of XAI methods, with $M_2=$\{\textit{GradientSHAP, IntegratedGradients}\}, $M_3=$\{\textit{Saliency, LRP-${z^+}$, Input$\times$Gradient}\},$M_4=$\{\textit{Gradient, GradCAM, LRP-$\varepsilon$, Guided-Backprop}\}, $M_5=${\textit{Guided-Backprop, GradientSHAP,  GradCAM, LRP-$\epsilon$, Saliency}\}}. The colour intensity denotes the task, i.e., the dataset and model combination.
    }
    \label{fig:average-mc}
    \vspace{-0.75em}
\end{figure}

As Figure \ref{fig:average-mc} illustrates, MPRT is generally outperformed by any of its variants (eMPRT or sMPRT) across the tested tasks and attribution sets $\{M_2, \cdots, M_5\}$ (see symbols over grouped bars). eMPRT and sMPRT consistently exhibit high reliability, substantiating its general applicability. 
These results are encouraging, as they suggest that the modifications made to MPRT have rendered the original metric more reliable. That said, Figure \ref{fig:average-mc} shows that no variant showcases perfect reliability, i.e., where $\overline{\text{MC}}=1$. This underlines the need for a continuously cautious application of XAI metrics, reflecting the broader challenge in XAI where ground truth explanation labels are often absent. Despite the performance advantage that eMPRT and sMPRT demonstrate, a contextualised (and not monolithic) metric adoption in XAI is strongly advocated. 

\section{Conclusion}

In this work, we proposed two extensions of the MPRT \cite{adebayo2018}, sMPRT and eMPRT, that address the caveats of pairwise similarity measures and layer randomisation order, 
which experimentally demonstrated improved reliability across various datasets- and model applications.
Note, however, that both proposed variations of the MPRT suffer from unique drawbacks as well, with sMPRT being computationally inefficient, hyperparameter-dependent and confounding the noise existing in the attributions, an arguably explanation method-specific property.
eMPRT, on the other hand, may possibly vary between different tasks, e.g., data-specific factors such as the size of the objects in a vision dataset. Each metric of evaluation has its own advantages and drawbacks, and we therefore emphasise the need for a broader evaluation, encompassing several properties to evaluate explanation quality.

Since both sMPRT and eMPRT employ fundamentally different mechanisms, a combination of both holds promise and will be subject to future work. Additionally, while we have started to explore the topic of layer-order for  randomisation (see the Supplements \ref{sec:layerorder_supplement}), this analysis is not complete, with especially bottom-up randomisation deserving further empirical validation.

\begin{ack}
This work was supported by the German Federal Ministry for Education and Research through project Explaining 4.0 (ref. 01IS200551);
BIFOLD (ref. 01IS18025A and ref. 01IS18037A); 
the Investitionsbank Berlin through BerDiBa (grant no. 10174498); the European Union’s Horizon Europe research and innovation programme (EU Horizon Europe) as grant TEMA (101093003);
and the European Union’s Horizon 2020 research and innovation programme (EU Horizon 2020) as grant iToBoS (965221).

\end{ack}

\bibliographystyle{IEEEtranEdited}
\bibliography{refs}

\newpage

\section{Supplementary Material}

The Supplementary material first describes experimental hyperparameters used in the main paper in detail and then discusses several additional experimental results.

\subsection{Experimental Setup}
\label{sec:experimental_setup_supplement}

\paragraph{Software Environment.} All experiments were implemented in python, using PyTorch \cite{Pytorch2019} for deep learning. zennit \cite{Anders2021Software} and captum \cite{Kokhlikyan2020Captum} packages were used to compute explanations. The Quantus \cite{hedstrom2022quantus} package was utilised to for evaluations with MPRT and extended to implement sMPRT and eMPRT. The MetaQuantus library was employed for meta-evaluation \cite{hedstrom2023metaquantus}.

\paragraph{Data.}
As shown by \cite{yona2021, kokhlikyan2021investigating}, MPRT can be sensitive to the choice of task. We therefore aim to cover a wide range of datasets in this work.
We use three image classification datasets in the experiments: ImageNet (ILSVRC2012 \cite{Krizhevsky2012Imagenet}), MNIST\cite{lecun2010mnist} and fMNIST \cite{fashionmnist2015}. For MNIST and fMNIST, we randomly sample $1000$ test samples and for ImageNet we use the first $1000$ samples of the validation set for experiments discussed in Figure \ref{fig:smprt} and randomly select $300$ test samples for experiments showcased in Figure \ref{fig:emprt} and Figures \ref{fig:average-mc}-\ref{fig:metaquantus-$M_5$}.

\paragraph{Models.}
The experiments are performed using different neural network models, including architectures such as LeNets \cite{lecun1998gradient} and ResNets \cite{he2015deep} and VGGs \cite{simonyan2014very},  
which contributes to the robustness of our findings. For MNIST and fMNIST, we train LeNets to an accuracy of $98.14$\% and $87.44$\% respectively. The training of all models is performed in a similar fashion; employing SGD optimisation with a standard cross-entropy loss, an initial learning rate of $0.001$ and momentum of $0.9$. All models are trained for $20$ epochs each. For ILSVRC-15 \cite{ILSVRC15} and ILSVRC2012 \cite{Krizhevsky2012Imagenet}, we use the ResNet-18 model \cite{he2015deep} and VGG-16 \cite{simonyan2014very} with pre-trained weights given the ImageNet dataset, accessible via \texttt{PyTorch} \cite{Pytorch2019}. 

\textbf{Explanation Methods and Preprocessing.} \label{preprocess}
We evaluated the following explanation methods using the MPRT variants:
\textit{Gradients} \cite{morch, baehrens}, \textit{Saliency} \cite{Simonyandeep}, \textit{Input$\times$Gradient} \cite{ShrikumarInputXG}, GradCAM \cite{selvaraju2019}, \textit{GradientSHAP} \cite{lundberg2017unified} with $5$ samples, \textit{SmoothGrad} \cite{smilkov2017smoothgrad} (with 20 noisy samples and a noise level of $0.1 \slash (x_\text{max}-x_\text{min})$ as implemented as default in zennit \cite{Anders2021Software}, \textit{Integrated Gradients} \cite{sundararajan2017axiomatic} (with 20 iterations and a baseline of zero), \textit{Guided Backpropagation} \cite{springenberg2014striving} and two distinct variations of \textit{Layer-wise Relevance Propagation} (LRP) \cite{bach2015pixel,montavon2017dtd}: application of the \textit{LRP-$\varepsilon$-rule} \cite{bach2015pixel} to all layers (called LRP-$\varepsilon$) and application of the \textit{LRP-${z^+}$-rule} \cite{montavon2017dtd} to all layers (called LRP-$z^+$). For random attribution (as a control variant) we sampled from a uniform distribution, i.e., $\hat{\ve}_i \sim \mathcal{U}(1, 0)$.

Since preprocessing can significantly affect the results of MPRT \cite{sundararajan2018}, we only take the absolute values for methods where the sign has no meaning in terms of feature importance (\textit{Saliency} and \textit{SmoothGrad}) and only consider the positive values for \textit{GradCAM}, as per the original paper \cite{selvaraju2019}. While the histogram entropy we used for eMPRT already normalises inputs implicitly, the \textit{SSIM} used to measure explanation difference in MPRT and sMPRT requires normalisation to make attributions at different scales before and after randomisation comparable. For this purpose, we utilised normalisation by the square root of the average second-moment estimate (cf. Supplement of \cite{bindershort}), as this normalisation does not impose as much additional variance as normalisation by, e.g., maximum value would. The normalisation is defined as follows:
\begin{equation}
\label{eq:normalise-eq}
   \text{norm}(\ve) = \frac{{\ve}_{h,w}}{\left(\frac{1}{HW}\sum_{h',w'}{\ve}_{h',w'}^2\right)^{1/2} }~,
\end{equation}
where $\hat{\ve}_{h,w}$ is the value of the explanation map at pixel location ($h$, $w$) and $H$, $W$ denote the height and width, respectively\footnote{This normalisation technique guarantees that the mean squared distance of each attribution score to zero equals one. As the procedure does not constrain attributions to a fixed range, it is not designed for visual representation but aims to preserve a measure useful for comparing the distances between different explanation methods.}. 

\paragraph{MPRT Hyperparameters.}
In the original paper \cite{adebayo2018}, different similarity measures are used e.g.,  \textit{Structural Similarity Index} (SSIM) \cite{nilsson2020understanding}, \textit{Spearman Rank Correlation} and \textit{HOG}. We employ \textit{SSIM} \cite{nilsson2020understanding}, motivated by its widespread adoption in existing literature \cite{adebayo2018,sixt2019,bindershort}.

\paragraph{sMPRT Hyperparameters.}
For the experiments with sMPRT, we employ \textit{SSIM} \cite{nilsson2020understanding} to measure explanation similarity. For generating noise ($\sigma$), we used an adaptive standard deviation of $0.2 \slash (x_\text{max}-x_\text{min})$, following the original heuristic presented in \cite{smilkov2017smoothgrad}. For the number of noise samples ($N$) to average over, we investigated the following values: $1, 20, 50, 300$. Evaluation for those values is shown in Supplement \ref{sec:smprt_supplement}. While a larger $N$ generally improves the denoised explanation estimate, it also increases computation time significantly. For Section \ref{sec:smprt}, we thus chose $N=50$ as our estimate with acceptable runtime.

\paragraph{eMPRT Hyperparameters.} \label{emprt_hyper}


In our experiments, for Equation \ref{eq:hist_entropy}, we empirically set the bin count $B=100$, which demonstrated robust performance across diverse experiments and explanation methods. Although various statistical rules like \textit{Freedman-Diaconis'} \cite{freedman1981histogram} and \textit{Scotts'} \cite{scott1979optimal} exist for optimising $B$, initial tests indicated inconsistency in their performance, largely due to their assumption of data normality, which is generally not applicable to attribution data.

To measure model complexity, we computed the Shannon entropy of the model output (post-softmax) probabilities as follows:
\begin{equation} \label{shannon-entropy}
\xi(f_L(\vx)) = -\sum_{i=1}^{|L|} p(x_i) \log_2(p(x_i))
\end{equation}

\paragraph{Meta-Evaluation Details.}
\label{sec:benchmarking_details}

To meta-evaluate the metrics, i.e.,  capture their metric performance characteristics in terms of statistical reliability, we employed the MetaQuantus library \cite{hedstrom2022quantus}. For this, we measured both their resilience to noise ($NR$) and their reactivity to adversary ($AR$) by inducing controlled perturbations, where we applied minor and disruptive perturbations to the input- and model spaces, respectively\footnote{We applied i.i.d additive uniform noise such that $\hat{\vx}_i = \vx + \bm{\delta}_i \ \text{with} \ \bm{\delta}_i \sim {\mathcal{U}}(\alpha, \beta)$ for the Input Perturbation Test and applied multiplicative Gaussian noise to all weights of the network, i.e., $\hat{\bm{\theta}}_i = \bm{\theta} \cdot \bm{\nu}_i \ \text{with} \ \bm{\nu}_i \sim\mathcal{N}(\bm{\mu}, \bm{\Sigma})$ for the Model Perturbation Test. The hyperparameters $\alpha, \beta, \mu, \Sigma$ were set according to the original publication \cite{hedstrom2023metaquantus}.}. Here, we computed the intra- ($\mathbf{IAC}$) and inter-consistency ($\mathbf{IEC}$) scores, which include measuring the similarity in score distributions and ranking of different explanation methods (post-perturbation), respectively. Each metric received a summarising meta-consistency (MC) score with $\text{MC}\in [0, 1]$:
\begin{equation}\label{eq:meta-eval-vector}
    \mathbf{MC} = \left(\frac{1}{|\mathbf{m}^{*}|}\right){\mathbf{m}^{*}}^T\mathbf{m} \quad \text{where} \quad \mathbf{m} = \begin{bmatrix} \mathbf{IAC}_{NR}\\ \mathbf{IAC}_{AR}\\ \mathbf{IEC}_{NR}\\     \mathbf{IEC}_{AR} \end{bmatrix},
 \end{equation}
with $\vm^{*} = \mathbb{R}^4$
represents an optimally performing quality estimator
as defined by the all-one indicator vector. For exact definitions for the contents of the meta-evaluation vector $\mathbf{m}$, we refer to the original publication \cite{hedstrom2023metaquantus}.

\paragraph{Meta-Evaluation Hyperparameters.}
\label{sec:benchmarking_details}

For empirical assessment, we utilised the pre-existing tests available in the MetaQuantus library \cite{hedstrom2022quantus} and their associated hyperparameters. The existing MetaQuatus tests are accessible via the GitHub repository\footnote{Code at: \url{https://github.com/annahedstroem/MetaQuantus/}.}. All metrics have been implemented in Quantus \cite{hedstrom2022quantus}. We executed these metrics over $K=5$ perturbations, spanning $3$ iterations with the test configurations as outlined in the notebook\footnote{See hyperparameter settings at notebook: \url{https://github.com/annahedstroem/MetaQuantus/blob/main/tutorials/Tutorial-Getting-Started-with-MetaQuantus.ipynb}}. To ensure a fair comparison among the metrics, we adhered to uniform hyperparameter settings, as specified by the normalisation formula in Equation \ref{eq:normalise-eq}. Given that the MetaQuantus library necessitates each metric to yield a single quality estimate $\hat{q} \in \mathbb{R}$ per sample, we calculated the correlation coefficient for the fully randomised models in both MPRT and sMPRT, which is anticipated to enhance their performance. 

\subsection{Effect of Layer Randomisation Order on MPRT Results}
\label{sec:layerorder_supplement}

\begin{figure}[!b]
    \centering
    \includegraphics[width=0.95\linewidth]{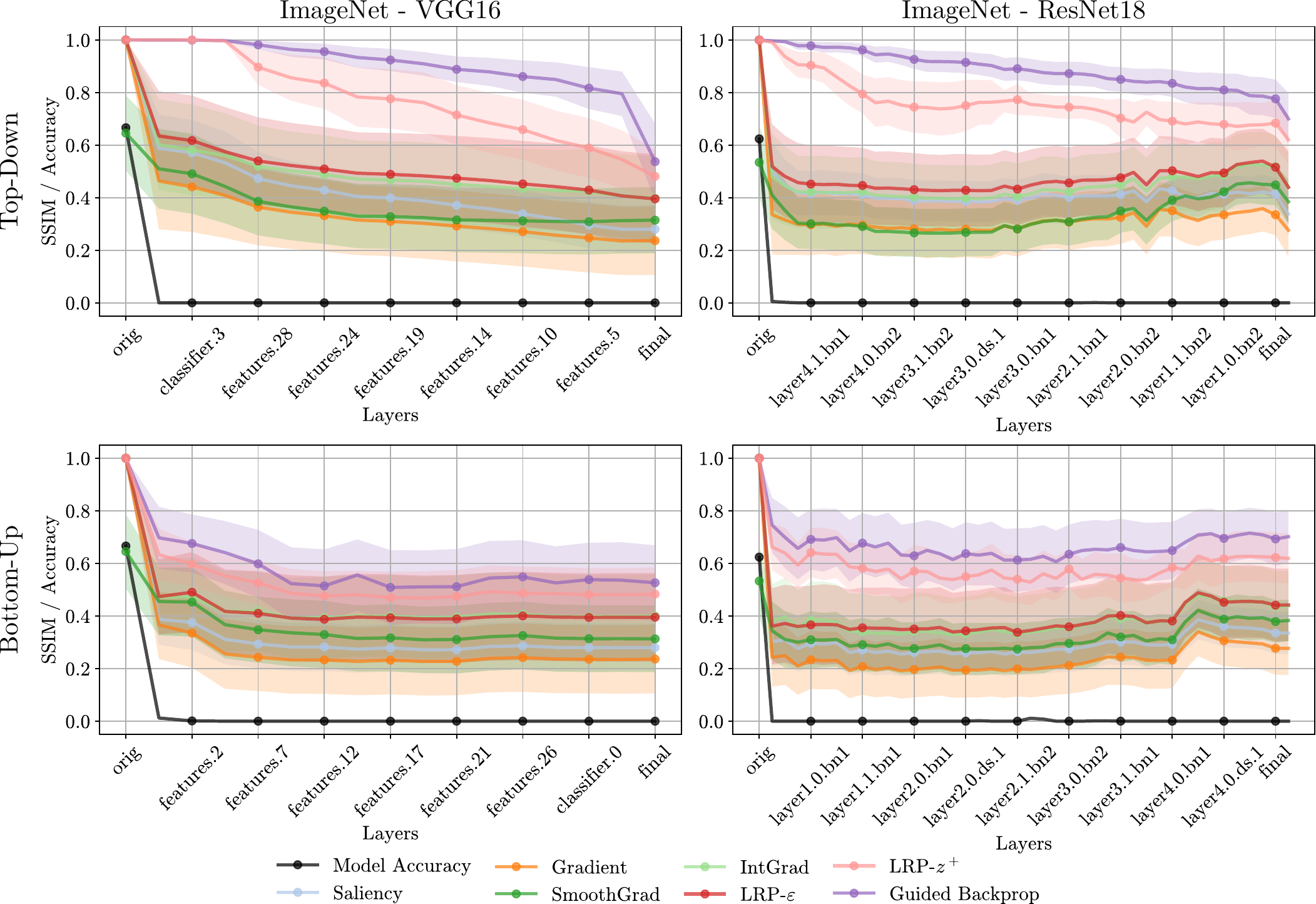}
    \captionsetup{font=footnotesize}
    \caption{Effect of layer randomisation order on MPRT results for ImageNet, using VGG16 \emph{(left)} and ResNet18 \emph{(right}). The \emph{black line} shows how model accuracy changes with randomisation, \emph{other lines} show the \textit{SSIM} measured by MPRT for different explanation methods. \emph{orig} and \emph{final} denote the unrandomised and fully randomised model states, respectively. \emph{(Top)} Top-down randomisation, starting with the output layer, as performed by \cite{adebayo2018}. \emph{(Bottom)} Bottom-up randomisation, starting with the lowest layer, as suggested by \cite{bindershort}. Model accuracy degrades with the same speed for both randomisation orders. However, there is an significant discrepancy in explanation method \textit{SSIM} scores: differences between methods seem much more severe for top-down than bottom-up randomisation. }
    \label{fig:mprt_layerorder}
\end{figure}

To avoid the preservation of significant portions of the forward pass that occurs with the latter (cf. and \ref{issues} (b) and \cite{bindershort}), \cite{bindershort} proposes to randomise layers in bottom-up order instead of top-down. Since bottom-up layer-order randomises the lowest layers first, such preservation cannot occur.

Given the use of bottom-up randomisation in sMPRT, and its current lack of empirical validation, we briefly examine its impact on MPRT in the following. 
Note that a better option may be to avoid progressive layer randomisation altogether, simply comparing the unrandomised state with the fully randomised state (see Equation \ref{eq:q_emprt}). 

Figure \ref{fig:mprt_layerorder} compares MPRT rankings for top-down \emph{(top)} and bottom-up \emph{(bottom)} randomisation order. Note that in contrast to the seminal work \cite{adebayo2018}, we apply a different preprocessing to the explanations where we normalise by the square root of the average second-moment estimate \cite{bindershort} (also cf. discussion in Section \ref{issues}, caveat (a)). We also take absolute values only for \textit{Saliency} and \textit{SmoothGrad}. 

First, we validate that the model is sufficiently disturbed by both orderings by also tracking the model accuracy. This is indeed the case for all four settings (see \emph{black} lines), where the accuracy drops to random for the ImageNet experiments (around $0.001$) immediately after the respective first layer is randomised. MPRT results are strongly affected by the randomisation order, however---not only do all explanation methods perform better with bottom-up randomisation, but since their \textit{SSIM} drops more steeply with randomisation, the differences in \textit{SSIM} score between methods significantly diminish. Especially, methods that appeared almost invariant to model parameters with top-down randomisation (\textit{LRP-$z^+$} and \textit{Guided Backpropagation}) show significantly more faithfulness to the model with bottom-up randomisation. 

\begin{figure}[!h]
    \centering
    \includegraphics[width=0.95\linewidth]{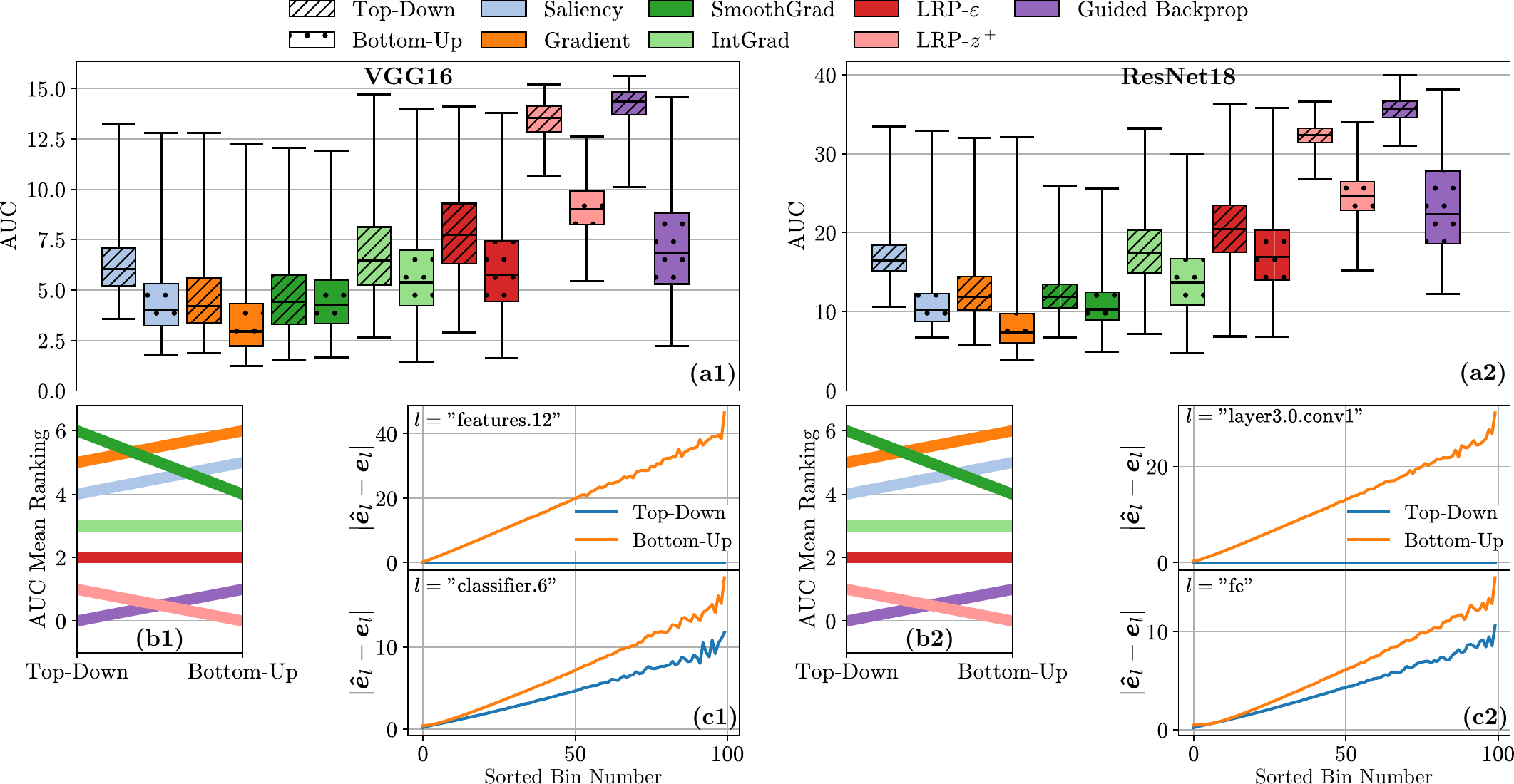}
    \captionsetup{font=footnotesize}
    \caption{Effect of layer randomisation order on MPRT results. Evaluation was performed for VGG16 \emph{(left)} and ResNet18 \emph{(right)} models, on the ImageNet dataset. \emph{a1 and a2}: Area under MPRT curves (cf. Figure \ref{fig:mprt_layerorder}) for various XAI methods and both layer orders. Percentile (0, 25, 50, 75, 100) statistics are visualised. \emph{b1 and b2}: Relative ranking changes between XAI methods from top-down to bottom-up order. Note that a lower score in (a) corresponds to a higher ranking in (b). \emph{(c1 and c2)}: Values of intermediate explanation $\ve$ (cf. \cite{bindershort}), assigned to 100 bins according to their absolute value (x-axis) vs. the average amount of change with randomisation for values in each bin (y-axis).} 
    \label{fig:mprt_layerorder_2}
\end{figure}

The above observations are confirmed by Figure \ref{fig:mprt_layerorder_2} \emph{(a1) and (a2)}: Explanation methods perform better under the MPRT with bottom-up randomisation (i.e., the dotted hatch bars are lower than the striped bars). While this does not translate to the categorical rankings between explanation methods  \emph{(b1) and (b2)}, which seem only slightly affected by randomisation order, keep in mind that there are also other interfering effects to consider, such as the sensitivity of \textit{SSIM} to noise \cite{bindershort}, as described in Section \ref{issues}, issue ({c}). These effects are likely to confound the bottom-up results as well and are addressed separately by the sMPRT and eMPRT metrics proposed in this work.

To understand why bottom-up randomisation leads to generally better MPRT performance, we inspect the (LRP-$\varepsilon$) explanations $\ve_l$ of the intermediate activations at layer $l$ ---i.e, $\ve_l$ explains the output of layer $l$ as opposed to the whole model (which would correspond to output layer $L$). The values of the original explanation $\ve_l$ are sorted into 100 equidistant bins according to their magnitude. We then randomise the model bottom-up or top-down, to compute randomised explanations $\hat{\ve}_l$. For bottom-up and top-down order, we randomise only the first and last parameterised layer of the model, respectively. 

For each bin, we evaluate how much the explanation values change on average under each randomisation (i.e., we compute the average $|\hat{\ve}_l-\ve_l|$ for each bin). Figure \ref{fig:mprt_layerorder_2} \emph{(c1) and (c2)} shows the results of this analysis for two different layers $l$---for each model, an intermediate layer \emph{(top)} in addition to the output layer \emph{(bottom)}.  Here, we observe that bottom-up randomisation (orange line) affects both the intermediate layer explanation and the last layer explanation, while top-down randomisation (blue line) only affects the latter (although significantly less than bottom-up randomisation). Bottom-up randomisation seems to have a stronger impact on the model's reasoning compared to top-down randomisation. For a robust MPRT, it is necessary to randomise the model output as fully as possible---because only then faithful explanations can be expected to change significantly \cite{bindershort}. In this aspect, bottom-up randomisation seems to be a better choice compared to top-down randomisation.
 
\subsection{sMPRT: Additional Experiments}
\label{sec:smprt_supplement}

In the following, we investigate the effect of the number of perturbed samples $N$ on sMPRT results. To make the curves obtained from sMPRT (cf. Figure \ref{fig:smprt}) comparable across different $N$, we condense them to a singular value by computing the area below each curve (denoted as AUC). Lower AUC indicates more faithfulness towards model parameters. 

\begin{figure}[!t]
    \centering
\includegraphics[width=0.95\linewidth]{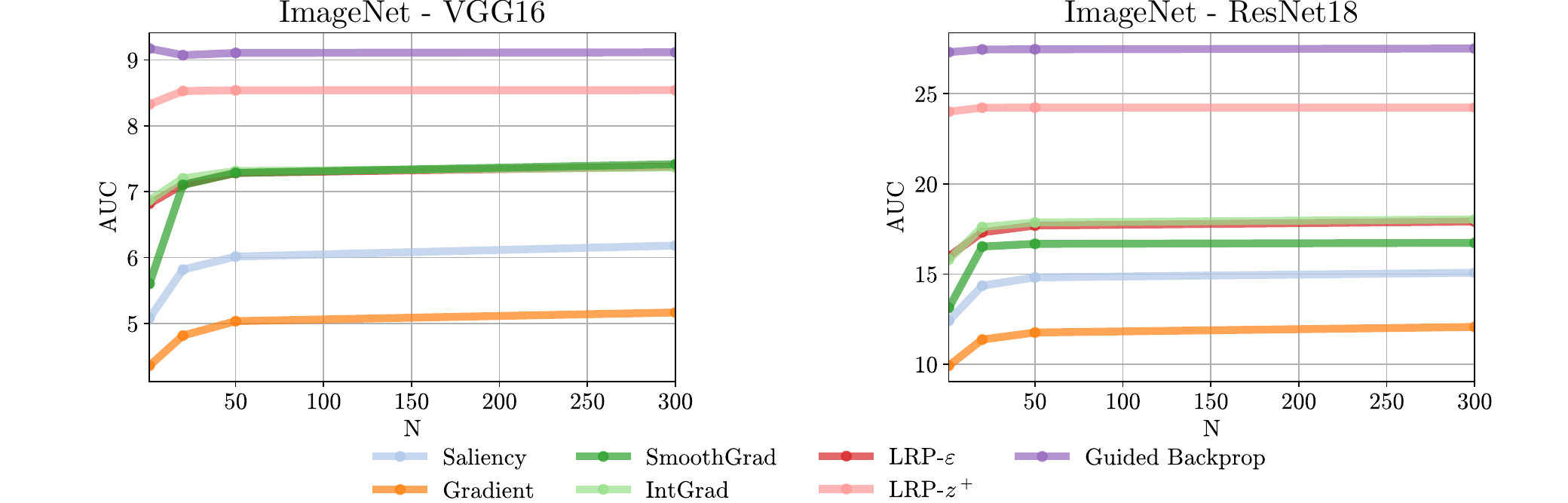}
    \captionsetup{font=footnotesize}
    \caption{Effect of number of perturbed samples $N$ on sMPRT results for VGG16 \emph{(left)} and ResNet18 \emph{(right)} on ImageNet data. The plots indicate how the area under the mean sMPRT curve (AUC) changes with $N$ for different explanation methods (i.e. the area under the curves as shown in Figure \ref{fig:smprt}). Up to $N=50$, there seems to be a steep change in AUC, especially for gradient-based methods. After that, the AUC curves flatten out, indicating a converged estimate of the denoised samples.}
    \label{fig:smprt_n}
\end{figure}

Figure \ref{fig:smprt_n} shows how these AUC values progress with increasing $N$. Especially gradient-based methods seem to vary significantly with $N$. However, the AUC seems to converge for large $N$, indicating that the estimation of the denoised samples also converges. While slight variations can be observed up to the last investigated value, $N=300$, AUC does not change much after $N=50$. For the sake of efficient evaluation, we therefore recommend to use $N=50$ to obtain a decent estimate of the correct ranking, albeit a larger $N$ will always yield a better estimation.

\subsection{eMPRT: Additional Experiments}
\label{sec:emprt_supplement}
\begin{figure}[!b]
    \centering
    \includegraphics[width=0.65\linewidth]{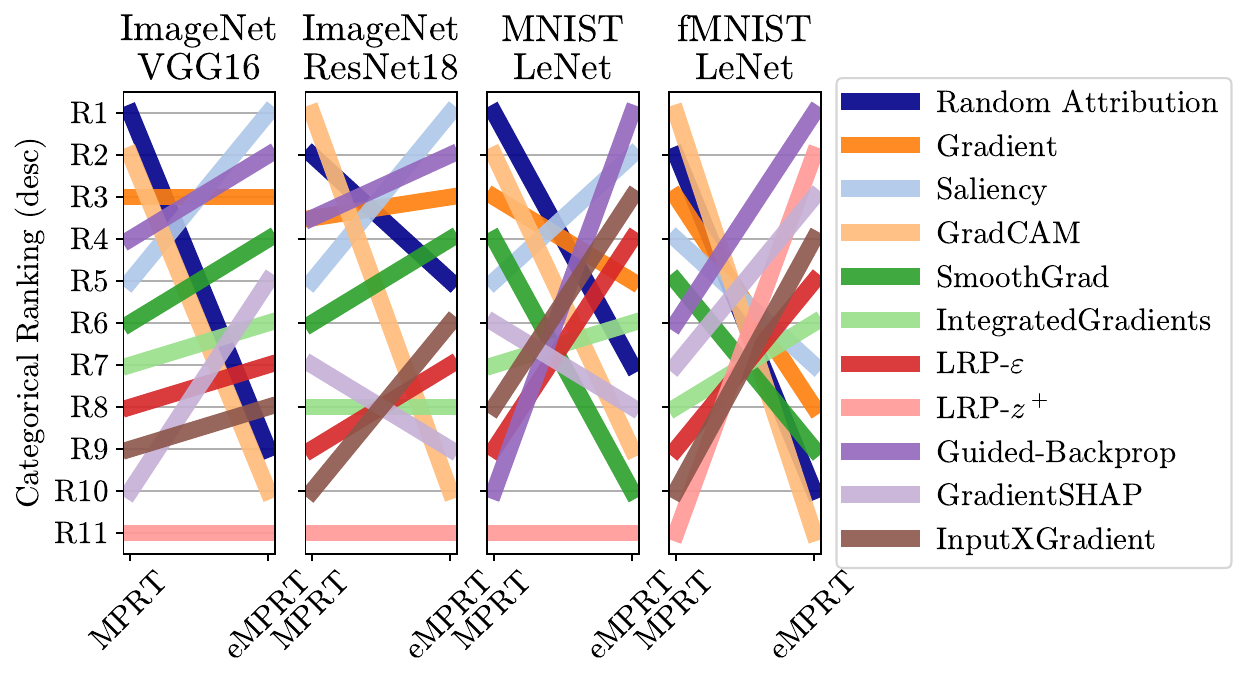}
    \captionsetup{font=footnotesize}
    \caption{Relative ranking of ten attribution methods and a random attribution using MPRT and eMPRT evaluation. The rankings, consistent with findings from \cite{hedstrom2023metaquantus}, vary significantly between the two metrics.}
    \label{fig:emprt-vs-mprt-ranking}
\end{figure}

In Figure \ref{fig:emprt-vs-mprt-ranking}, the categorical rankings derived from the evaluation metrics MPRT and eMPRT are visualised for ten distinct attribution methods, as well as a random attribution. Rankings are organised in descending order, where $R1$ denotes the best performance and 
$R11$ is the worst score. The findings corroborate those of \cite{hedstrom2023metaquantus}, demonstrating considerable variability in explanation rankings between the two metrics. In contrast to the original publication \cite{adebayo2018}, which asserted \textit{Guided Backpropagation} as inferior to gradient-based methods such as \textit{Gradient} and \textit{SmoothGrad}, our eMPRT evaluation reveals a contrasting ranking, even advancing \textit{Guided Backpropagation} above its gradient-based counterparts. The random attribution---serving as a theoretical lower bound for explanation quality--consistently receives lower scores for eMPRT compared to MPRT, signifying an advantage of our proposed metric.

\subsection{Benchmarking: Additional Details} \label{sec:benchmarking-extra-results}

In this section, we present a detailed breakdown of the benchmarking results. First, we present the numerical version of Figure \ref{fig:average-mc}, presented in the main manuscript. Here, $M_2=$\{\textit{GradientSHAP, IntegratedGradients}\}, $M_3=$\{\textit{Saliency, LRP-${z^+}$, Input$\times$Gradient}\},$M_4=$\{\textit{Gradient, GradCAM, LRP-Eps, Guided-Backprop}\}, $M_5=${\textit{Guided-Backprop, GradientSHAP,  GradCAM, LRP-$\epsilon$, Saliency}\}}. 

\begin{table}[h!]
    \centering
    \captionsetup{font=footnotesize}
    \caption{Meta-evaluation benchmarking results, aggregated over $3$ iterations with $K=5$. The top-performing metric in each setting is highlighted in bold. If it outperforms the next-best-performing metric by at least one standard deviation, it is \underline{underlined}. Higher values are preferred.}
    \label{table-benchmarking}
    \footnotesize
\begin{tabular}{ccccc}

\toprule
          \textbf{Setting} &                  \textbf{XAI Methods} &             \textbf{eMPRT} ($\uparrow$) &             \textbf{sMPRT} ($\uparrow$) &              \textbf{MPRT} ($\uparrow$) \\
\midrule
\multirow{4}{*}{\textit{fMNIST LeNet}} &              \textit{$M_2$} & \textbf{0.625 $\pm$ 0.002} & 0.613 $\pm$ 0.119 & 0.594 $\pm$ 0.088 \\
\cline{3-5}
 &     \textit{$M_3$} & \textbf{\underline{0.717 $\pm$ 0.012}} &  0.610 $\pm$ 0.083 & 0.653 $\pm$ 0.058 \\
 \cline{3-5}
 &     \textit{$M_4$} & \underline{\textbf{0.677 $\pm$ 0.091}} & 0.643 $\pm$ 0.005 & 0.617 $\pm$ 0.033 \\
 \cline{3-5}
 &     \textit{$M_5$} & 0.626 $\pm$ 0.064 &  \underline{\textbf{0.680 $\pm$ 0.023}} & 0.647 $\pm$ 0.057 \\
 \midrule
\multirow{4}{*}{\textit{MNIST LeNet}} &              \textit{$M_2$} &  \underline{\textbf{0.690 $\pm$ 0.051}} & 0.608 $\pm$ 0.111 &   0.550 $\pm$ 0.030 \\
\cline{3-5}
 &     \textit{$M_3$} & \underline{\textbf{0.740 $\pm$ 0.036}} &  0.580 $\pm$ 0.006 &   0.600 $\pm$ 0.048 \\
 &     \textit{$M_4$} & \underline{\textbf{0.653 $\pm$ 0.046}} & 0.555 $\pm$ 0.015 & 0.539 $\pm$ 0.019 \\
 \cline{3-5}
 &     \textit{$M_5$} & 0.618 $\pm$ 0.012 & \textbf{0.623 $\pm$ 0.004} & 0.597 $\pm$ 0.012 \\
  \midrule
\multirow{4}{*}{\textit{ImageNet ResNet18}} &              \textit{$M_2$} & 0.577 $\pm$ 0.047 & \textbf{0.608 $\pm$ 0.042} & 0.576 $\pm$ 0.045 \\
\cline{3-5}
 &     \textit{$M_3$} &   \textbf{0.709  $\pm$ 0.047} &  0.658 $\pm$ 0.054 & {{0.691 $\pm$ 0.045}} \\
 \cline{3-5}
 &     \textit{$M_4$} & 0.646 $\pm$ 0.052 &  \textbf{0.678 $\pm$ 0.050} & 0.673 $\pm$ 0.057 \\
 \cline{3-5}
 &     \textit{$M_5$} & 0.628 $\pm$ 0.023 &  \textbf{0.651 $\pm$ 0.050} & 0.639 $\pm$ 0.039 \\
 \midrule
\multirow{4}{*}{\textit{ImageNet VGG16}} &              \textit{$M_2$} & \textbf{0.548 $\pm$ 0.032} & 0.533 $\pm$ 0.06 & 0.523 $\pm$ 0.043 \\
\cline{3-5}
 &     \textit{$M_3$} & \textbf{0.651 $\pm$ 0.047} & 0.649 $\pm$ 0.051 &   0.650 $\pm$ 0.050 \\
 \cline{3-5}
 &     \textit{$M_4$} &  0.630 $\pm$ 0.041 & {\textbf{0.688 $\pm$ 0.052}} & 0.637 $\pm$ 0.053 \\
 \cline{3-5}
 &     \textit{$M_5$} &  \textbf{0.610 $\pm$ 0.039} & 0.613 $\pm$ 0.041 & 0.597 $\pm$ 0.043 \\
\bottomrule
\end{tabular}
\end{table}

In the following Figures \ref{fig:metaquantus-$M_2$}-\ref{fig:metaquantus-$M_5$}, we show the different area graphs which each contains the entries of the meta-evaluation vector (set as coordinates on a 2D plane).

\begin{figure}[!h]
    \centering
    \includegraphics[width=0.75\linewidth]{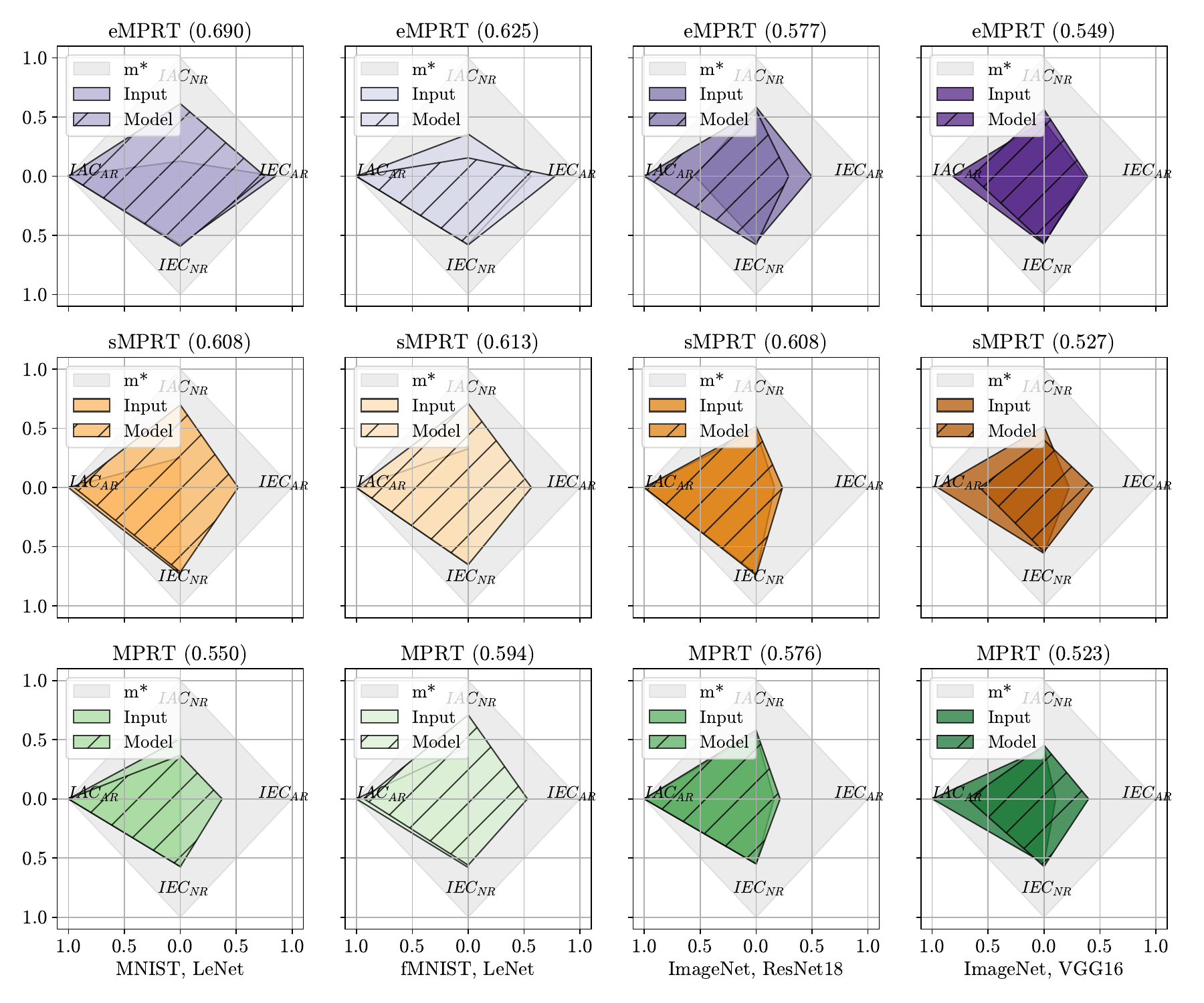}
    \caption{A graphical representation of the benchmarking results for $M=2$ for MPRT, sMPRT and eMPRT, aggregated over $3$ iterations with $K=5$. The grey area indicates the area of an optimally performing estimator, i.e., $\mathbf{m}^{*} = \mathbb{R}^4$. The MC score (indicated in brackets) is averaged over MPT and IPT. Higher values are preferred.} 
    \label{fig:metaquantus-$M_2$}
\end{figure}

By inspecting the coloured areas of the respective estimators in terms of their size and shape, we can deduce the overall performance of both failure modes (N$R$, $AR$). Here, larger coloured areas imply better performance on the different scoring criteria and the grey area indicates the area of an optimally performing quality estimator, i.e., $\vm^{*} = \mathbb{R}^4$. The shades of the different colours, i.e., purple, orange and green indicate the datasets and are used consistently across the Figures \ref{fig:metaquantus-$M_2$}-\ref{fig:metaquantus-$M_5$} (as well as Figure \ref{fig:average-mc} in the main manuscript). The columns show the tasks and the rows the different metrics: MPRT, sMPRT and eMPRT. With few exceptions, eMPRT and sMPRT demonstrate superior performance, on both the input- and the model perturbation tests, across tested tasks and subsets of attribution methods.

\begin{figure}[!t]
    \centering
    \includegraphics[width=0.75\linewidth]{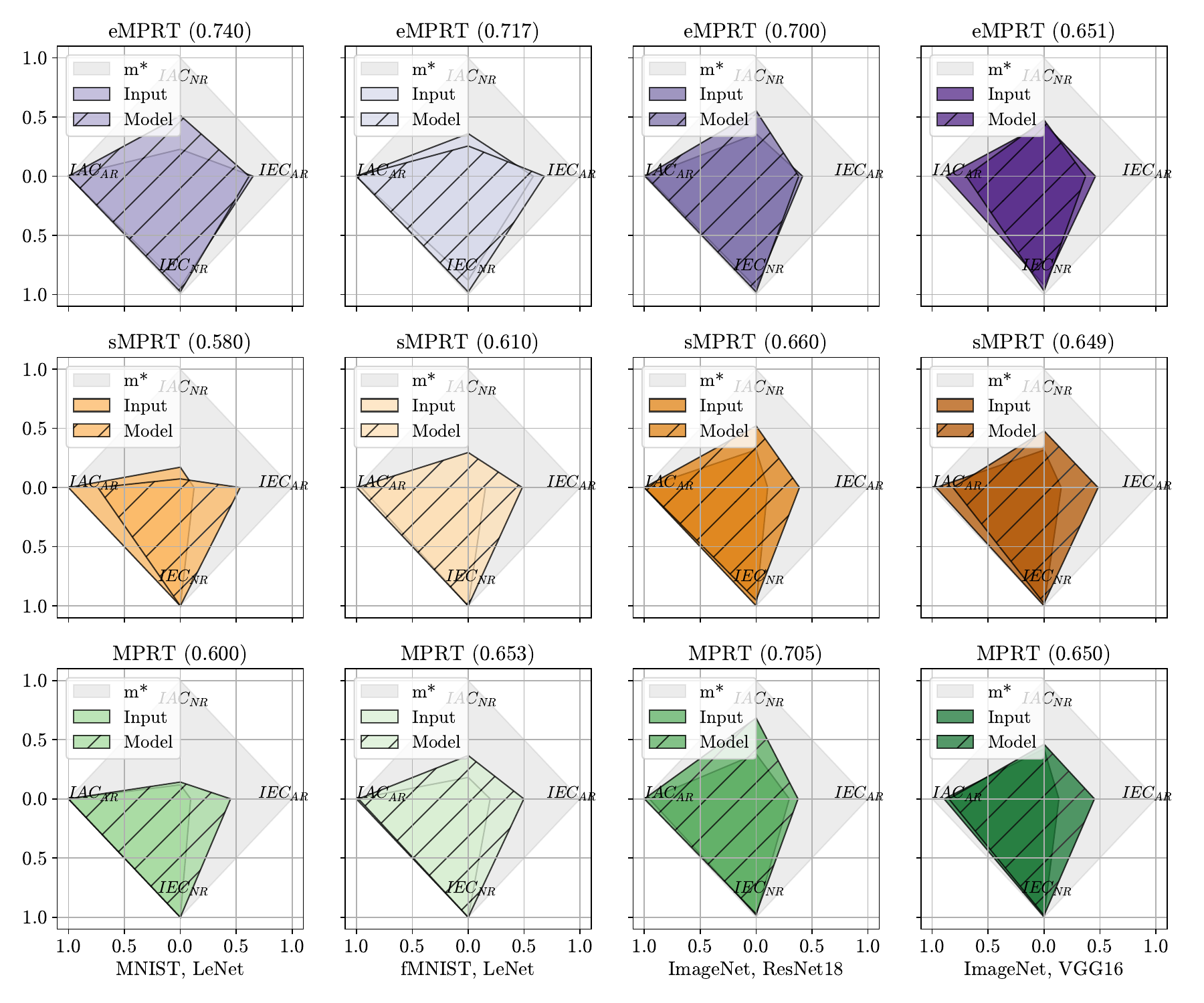}
    \caption{A graphical representation of the benchmarking results for $M=2$ for MPRT, sMPRT and eMPRT, aggregated over $3$ iterations with $K=5$. The grey area indicates the area of an optimally performing estimator, i.e., $\mathbf{m}^{*} = \mathbb{R}^4$. The MC score (indicated in brackets) is averaged over MPT and IPT. Higher values are preferred.} 
    \label{fig:metaquantus-$M_3$}
\end{figure}

\begin{figure}[!t]
    \centering
    \includegraphics[width=0.75\linewidth]{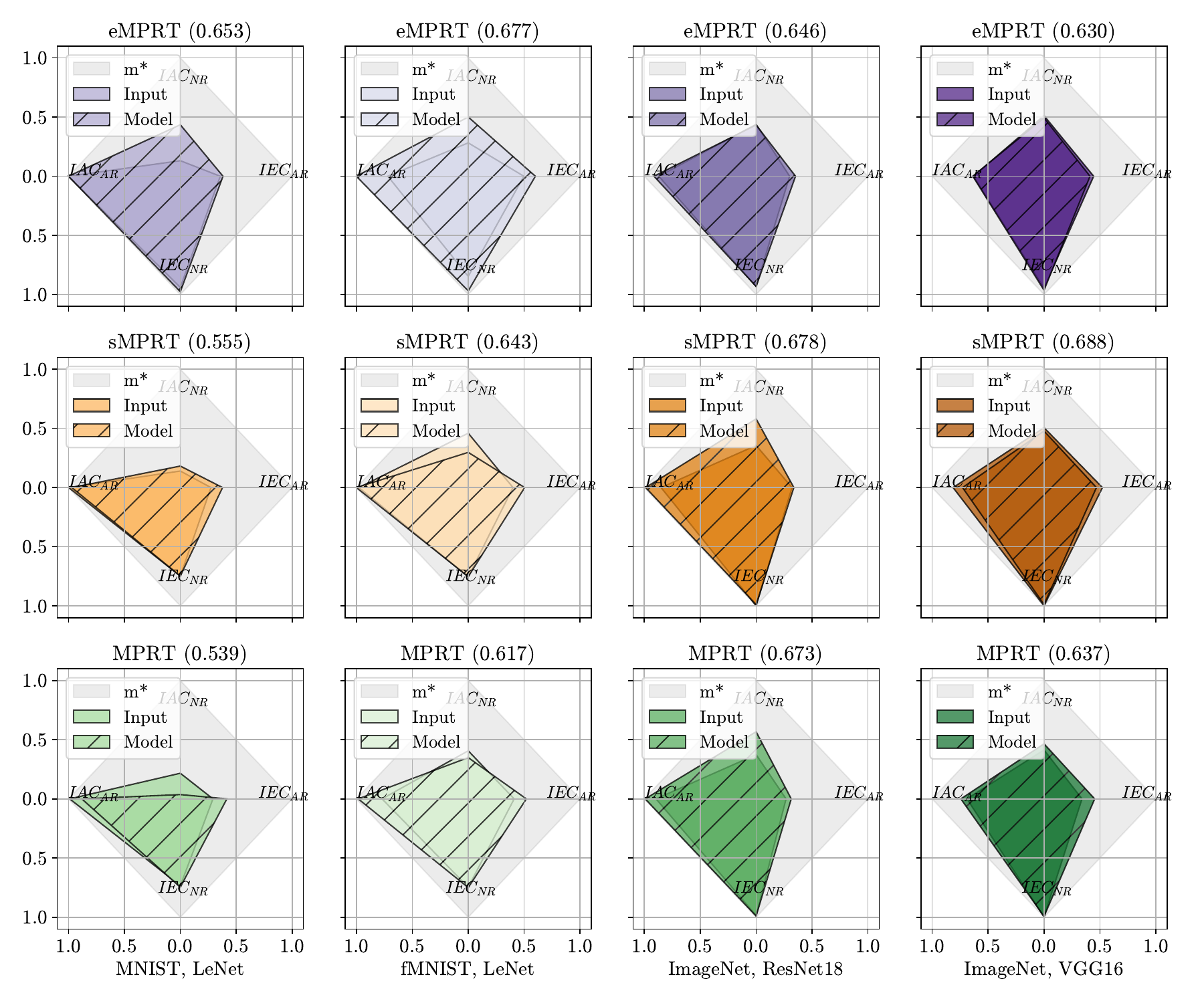}
    \caption{A graphical representation of the benchmarking results for $M=2$ for MPRT, sMPRT and eMPRT, aggregated over $3$ iterations with $K=5$. The grey area indicates the area of an optimally performing estimator, i.e., $\mathbf{m}^{*} = \mathbb{R}^4$. The MC score (indicated in brackets) is averaged over MPT and IPT. Higher values are preferred.} 
    \label{fig:metaquantus-$M_4$}
\end{figure}

\begin{figure}[!t]
    \centering
    \includegraphics[width=0.75\linewidth]{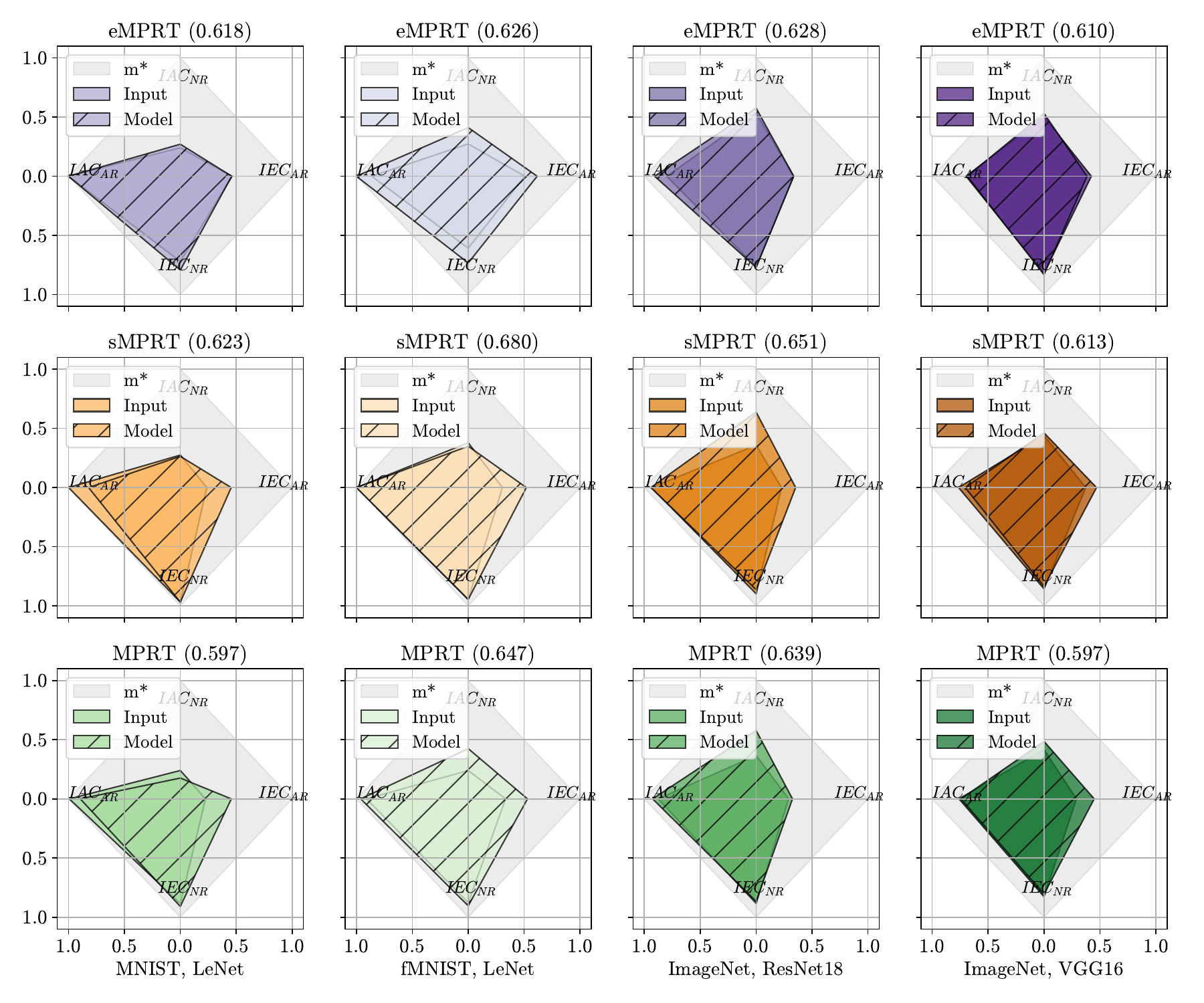}
    \caption{A graphical representation of the benchmarking results for $M=2$ for MPRT, sMPRT and eMPRT, aggregated over $3$ iterations with $K=5$. The grey area indicates the area of an optimally performing estimator, i.e., $\mathbf{m}^{*} = \mathbb{R}^4$. The MC score (indicated in brackets) is averaged over MPT and IPT. Higher values are preferred.} 
    \label{fig:metaquantus-$M_5$}
\end{figure}


\end{document}

%% file: math_commands.tex
\usepackage{amsmath}
\usepackage{amsthm} 
\usepackage{amssymb}
\usepackage{amsfonts}
\usepackage{bm}

\def\eqref#1{equation~\ref{#1}}

\def\1{\bm{1}}

\def\ve{{\bm{e}}}

\def\vm{{\bm{m}}}

\def\vx{{\bm{x}}}
\def\vy{{\bm{y}}}

\DeclareMathAlphabet{\mathsfit}{\encodingdefault}{\sfdefault}{m}{sl}
\SetMathAlphabet{\mathsfit}{bold}{\encodingdefault}{\sfdefault}{bx}{n}

\def\sF{{\mathbb{F}}}

\def\sO{{\mathbb{O}}}

\def\sX{{\mathbb{X}}}
\def\sY{{\mathbb{Y}}}

